\documentclass[runningheads]{llncs}
\usepackage{graphicx}
\usepackage{subfigure}
\usepackage{tikz}
\usepackage{comment}
\usepackage{amsmath,amssymb} % define this before the line numbering.
\usepackage{color}
\usepackage{booktabs}
\usepackage[accsupp]{axessibility}  % Improves PDF readability for those with disabilities.
\usepackage{hyperref}
\begin{document}
% \linenumbers
\pagestyle{headings}
\mainmatter
\def\ECCVSubNumber{8}  % Insert your submission number here

\title{Bag of Tricks for Out-of-Distribution Generalization} % Replace with your title

% % INITIAL SUBMISSION 
% %\begin{comment}
% \titlerunning{ECCV-22 submission ID \ECCVSubNumber} 
% \authorrunning{ECCV-22 submission ID \ECCVSubNumber} 
% \author{Anonymous ECCV submission}
% \institute{Paper ID \ECCVSubNumber}
% %\end{comment}
% %******************

% \begin{comment}
\author{Zining Chen\inst{1} \and
Weiqiu Wang\inst{1} \and
Zhicheng Zhao\inst{1,2} \and
Aidong Men\inst{1} \and
Hong Chen\inst{3}}
\authorrunning{Z. Chen et al.}
\institute{Beijing University of Posts and Telecommunications\and
Beijing Key Laboratory of Network System and Network Culture, China \and
China Mobile Research Institute
\email{\{chenzn,wangweiqiu,zhaozc,menad\}@bupt.edu.cn}\\
\email{chenhongyj@chinamobile.com}}
% \end{comment}
%******************
\maketitle

\begin{abstract}
% Recent research on out-of-distribution generalization increases the robustness and generalization ability of deep learning based models. To make contributions to this issue, nicochallenge-2022 provides NICO++, a large-scale dataset with diverse context information. However, most current algorithms are specifically designed for certain dataset and may not perform well on NICO++ dataset. In this paper, our team proposes bag of tricks for out-of-distribution generalization, aiming to provide effective methods for image recognition on domain generalization. We examine methods on network architecture, data augmentations, training and inference strategies, and achieve an excellent performance with the result of 88.16$\%$ on public test set and rank 1st in Track 1 of nicochallenge-2022.
Recently, out-of-distribution (OOD) generalization has attracted attention to the robustness and generalization ability of deep learning based models, and accordingly, many strategies have been made to address different aspects related to this issue. However, most existing algorithms for OOD generalization are complicated and specifically designed for certain dataset. To alleviate this problem, nicochallenge-2022 provides NICO++, a large-scale dataset with diverse context information. In this paper, based on systematic analysis of different schemes on NICO++ dataset, we propose a simple but effective learning framework via coupling bag of tricks, including multi-objective framework design, data augmentations, training and inference strategies. Our algorithm is memory-efficient and easily-equipped, without complicated modules and does not require for large pre-trained models. It achieves an excellent performance with Top-1 accuracy of 88.16$\%$ on public test set and 75.65$\%$ on private test set, and ranks $1^{st}$ in domain generalization task of nicochallenge-2022.

\end{abstract}

\section{Introduction}
\par
Deep learning based methods usually assume that data in training set and test set are independent and identically distributed (IID). However, in real world scenario, test data may have large distribution shifts to training data, leading to significant decrease on model performance. Thus, how to enable models to tackle data distribution shifts and better recognize out-of-distribution data is a topic of general interest nowadays. Nicochallenge-2022 is a part of ECCV-2022 which aims at facilitating the out-of-distribution generalization in visual recognition, searching for methods to increase model generalization ability, and track 1 mainly focuses on Common Context Generation (Domain Generalization, DG).
\par
Advancements in domain generalization arise from multiple aspects, such as feature learning, data processing and learning strategies. However, as distribution shifts vary between datasets, most of the existing methods have limitations on generalization ability. Especially NICO++ dataset is a large-scale dataset containing 60 classes in track 1, with hard samples including different contexts, multi-object and occlusion problems, etc. Therefore, large distribution shifts between current domain generalization datasets and NICO++ may worsen the effect of existing algorithms.
\par
% In this paper, we propose bag of tricks on track 1 of nicochallenge-2022, which consumes negligible computational resources in training and test stage, but makes a huge difference on model accuracy. First, we elaborately design a multi-task classifier structure and use multi-label to supervise, enabling models to learn coarse context of data and enrich semantic information. Then, we apply different data augmentations and training strategies to increase model robustness and generalization ability. Also, we use test-time augmentations to improve test performance and employ several vision models with different architectures to complement each other for model ensemble. Finally, we get a result of 88.16$\%$ with our method on public test set and rank 1st in phase 1.
In this paper, without designs of complicated modules, we systematically explore existing methods which improve the robustness and generalization ability of models. We conduct extensive experiments mainly on four aspects: multi-objective framework design, data augmentations, training and inference strategies. Specifically, we first compare different ways to capture coarse-grained information and adopt coarse-grained semantic labels as one of the objective in our proposed multi-objective framework. Secondly, we explore different data augmentations to increase the diversity of data to avoid overfitting. Then, we design a cyclic multi-scale training strategy, which introduces more variations into the training process to increase model generalization ability. And we find that enlarging input size is also helpful. Moreover, we merge logits of different scales to make multi-scale inference and design weighted Top-5 voting to ensemble different models. Finally, our end-to-end framework with bag of simple and effective tricks, as shown in Figure~\ref{fig:net}, gives out valuable practical guidelines to improve the robustness and generalization ability of deep learning models. Our solution achieves superior performance on both public and private test set of domain generalization task in nicochallenge-2022, with the result of 88.16$\%$ and 75.65$\%$ respectively, and ranks $1^{st}$ in both phases.

\section{Related Work}

\subsection{Domain Generalization}
Domain Generalization aims to enable models to generalize well on unknown-distributed target domains by training on source domains. Domain-invariant feature learning develops rapidly in past few years, IRM~\cite{arjovsky2019invariant} concentrates on learning an optimal classifier to be identical across different domains, CORAL~\cite{sun2016deep} aims at feature alignment by minimizing second-order statistics of source and target domain distributions. Data processing methods including data generation (e.g. Generative Adversarial Networks~\cite{goodfellow2014generative}) and data augmentation (e.g. Rotation) are simple and useful to increase the diversity of data, which is essential in domain generalization. Other strategies include Fish~\cite{shi2021gradient}, a multi-task learning strategy that consists the direction of descending gradient between different domains. StableNet~\cite{zhang2021deep} aims to extract essential features from different categories and remove irrelevant features and fake associations by using Random Fourier Feature. SWAD~\cite{cha2021swad} figures out that flat minima leads to smaller domain generalization gaps and suffers less from overfitting. Several self-supervised learning methods~\cite{kim2021selfreg}~\cite{feng2019self}~\cite{carlucci2019domain} are also proposed these years to effectively learn intrinsic image properties and extract domain-invariant representations. Although these methods make great progress on domain generalization, most of them are complicatedly designed and may only benefit on certain dataset.

\subsection{Fine-grained Classification}
Fine-grained Classification aims to recognize sub-classes under main classes. Difficulty mainly lies in finer granularity for small inter-class variances, large intra-class similarity and different image properties (e.g. angle of view, context and occlusion). Attention mechanisms are mainstream of fine-grained classification which aim at more discriminative foreground features and suppress irrelevant background information~\cite{fu2017look}~\cite{wang2017residual}~\cite{hu2018squeeze}. Also, network ensemble methods (e.g. Multiple granularity CNN~\cite{wang2015multiple}) including dividing classes into sub-classes or using multi-branch neural networks are proposed. Meanwhile, high-order fine-grained feature is another aspect, which Bilinear CNN~\cite{lin2015bilinear} uses second-order statistics to fuse context of different channels. However, as fine-grained based methods may have different effects between networks and several with high computational complexity, we only adopt light-weight ECA channel attention mechanism in eca-nfnet-l0 backbone network~\cite{brock2021high} and SE channel attention mechanism in efficientnet-b4 backbone network~\cite{tan2019efficientnet}.

\subsection{Generalization Ability}
Generalization Ability refers to the adaptability of models on unseen data, which is usually relevant to model overfitting in deep learning based approaches. Reduce model complexity can avoid model fitting into a parameter space only suitable for training set. For example, use models with less parameters and add regularization terms (e.g. L1 and L2 regularization) to limit the complexity of models~\cite{goodfellow2016regularization}. Diverse data distribution can also increase generalization ability by using abundant data for pre-training (e.g. Imagenet~\cite{deng2009imagenet}), applying data augmentation methods~\cite{shorten2019survey}, and using re-balancing strategies to virtually set different-distributed dataset~\cite{zhou2020bbn}. 

\section{Challenge Description}
\subsection{Dataset}
\begin{figure*}[htbp]
\centering
\subfigure[Multi-target.]{
\begin{minipage}[b]{1\linewidth}
\centering
\begin{minipage}[b]{0.30\linewidth}
\includegraphics[width=0.9\linewidth, height=0.9\linewidth]{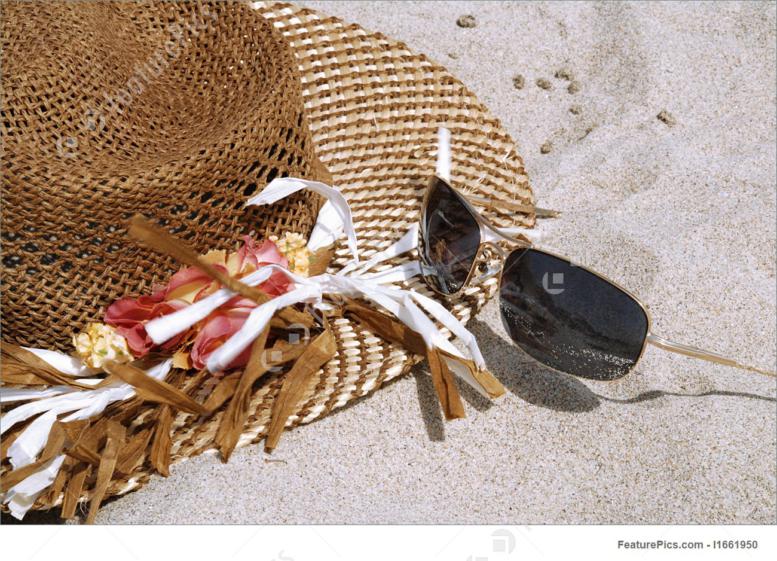}
\end{minipage}
\begin{minipage}[b]{0.30\linewidth}
\includegraphics[width=0.9\linewidth, height=0.9\linewidth]{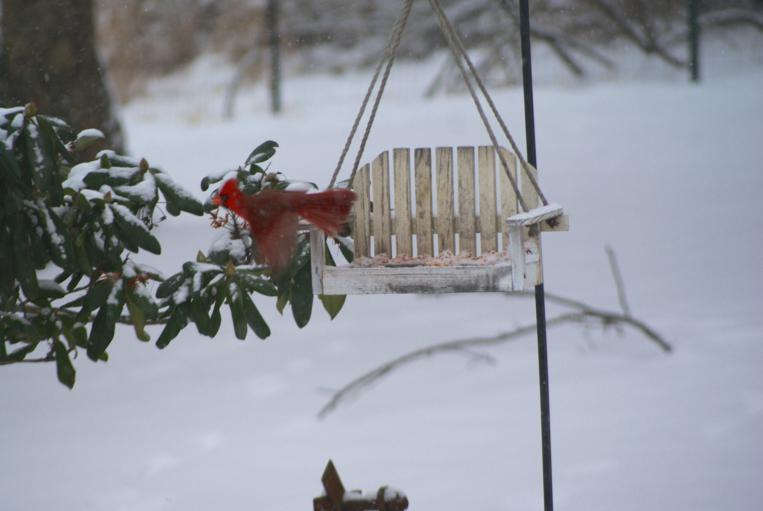}
\end{minipage}
\begin{minipage}[b]{0.30\linewidth}
\includegraphics[width=0.9\linewidth, height=0.9\linewidth]{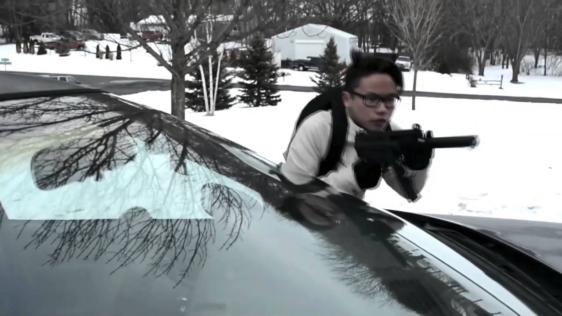}
\end{minipage}
\end{minipage}}
\subfigure[Occlusions.]{
\begin{minipage}[b]{1\linewidth}
\centering
\begin{minipage}[b]{0.30\linewidth}
\includegraphics[width=0.9\linewidth, height=0.9\linewidth]{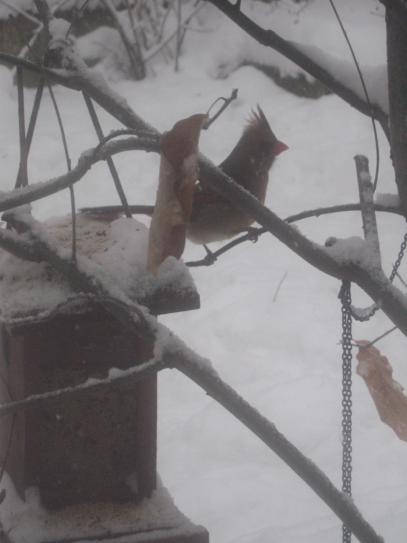}
\end{minipage}
\begin{minipage}[b]{0.30\linewidth}
\includegraphics[width=0.9\linewidth, height=0.9\linewidth]{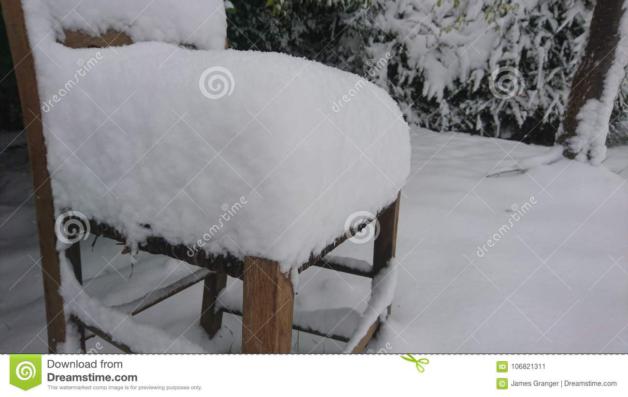}
\end{minipage}
\begin{minipage}[b]{0.30\linewidth}
\includegraphics[width=0.9\linewidth, height=0.9\linewidth]{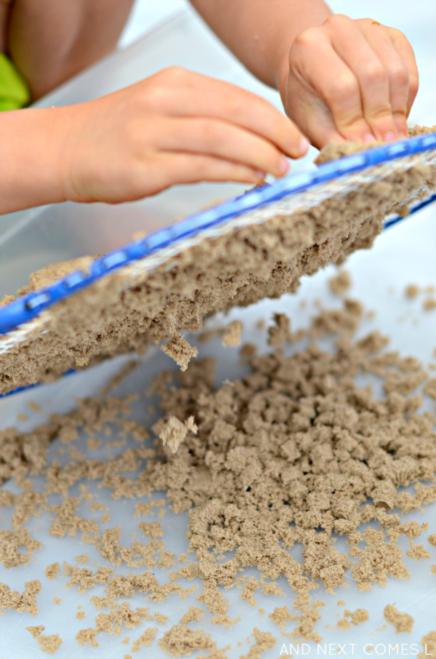}
\end{minipage}
\end{minipage}}
\subfigure[Angle of views.]{
\begin{minipage}[b]{1\linewidth}
\centering
\begin{minipage}[b]{0.30\linewidth}
\includegraphics[width=0.9\linewidth, height=0.9\linewidth]{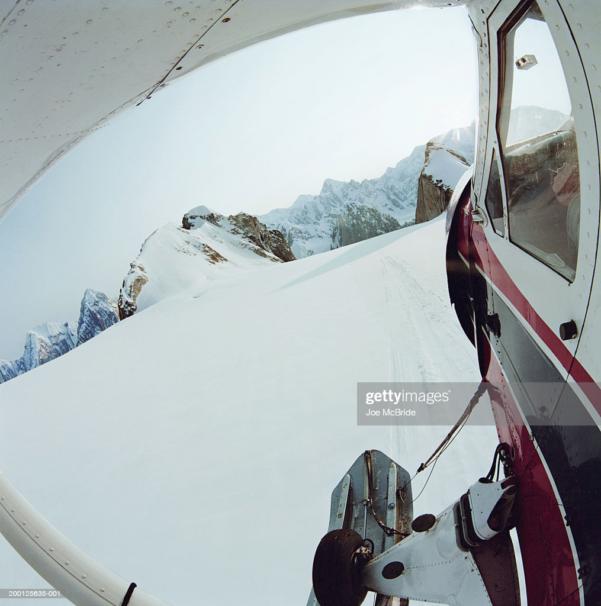}
\end{minipage}
\begin{minipage}[b]{0.30\linewidth}
\includegraphics[width=0.9\linewidth, height=0.9\linewidth]{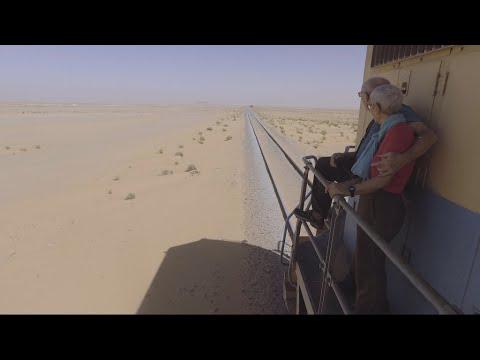}
\end{minipage}
\begin{minipage}[b]{0.30\linewidth}
\includegraphics[width=0.9\linewidth, height=0.9\linewidth]{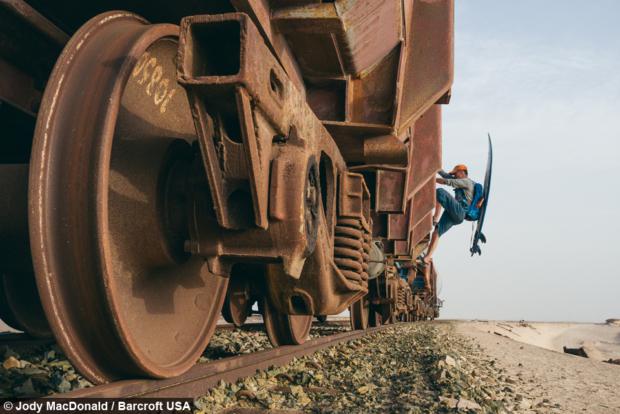}
\end{minipage}
\end{minipage}}
\caption{Hard samples with difficult image properties, such as multi-target, occlusions and angle of views, which are easily classified incorrectly for many models.}
\label{figure:1}
\end{figure*}

The data of domain generalization task in nicochallenge-2022 is from NICO++ dataset~\cite{zhang2022nico}, a novel domain generalization dataset consisting of 232.4k images for total 80 categories, including 10 common domains and 10 unique domains for each category. The data in domain generalization task is reorganized to 88,866 samples for training, 13,907 for public test and 35,920 for private test with 60 categories. While images from most domains are collected by searching a combination of a category name and a phrase extended from the domain name, there exists hard samples with multi-target, large occlusions and different angle of views, as shown in Figure~\ref{figure:1}.
\renewcommand{\floatpagefraction}{.9}
\subsection{Task}
Track 1 of nicochallenge-2022 is a common context generation competition on image recognition which aims at facilitating the OOD generalization in visual recognition, whose contexts of training and test data for all categories are aligned and domain labels for training data are available. This task is also known as domain generalization, to perform better generalization ability on unknown test data distribution. Specifically, its difficulty mainly lies in no access to target domains with different distributions during training phase. Thus, the key for this challenge is to improve the robustness and generalization ability of models based on images with diverse context and properties in NICO++ dataset. 
\section{Method}
\begin{figure}
\centering
\includegraphics[width=120mm]{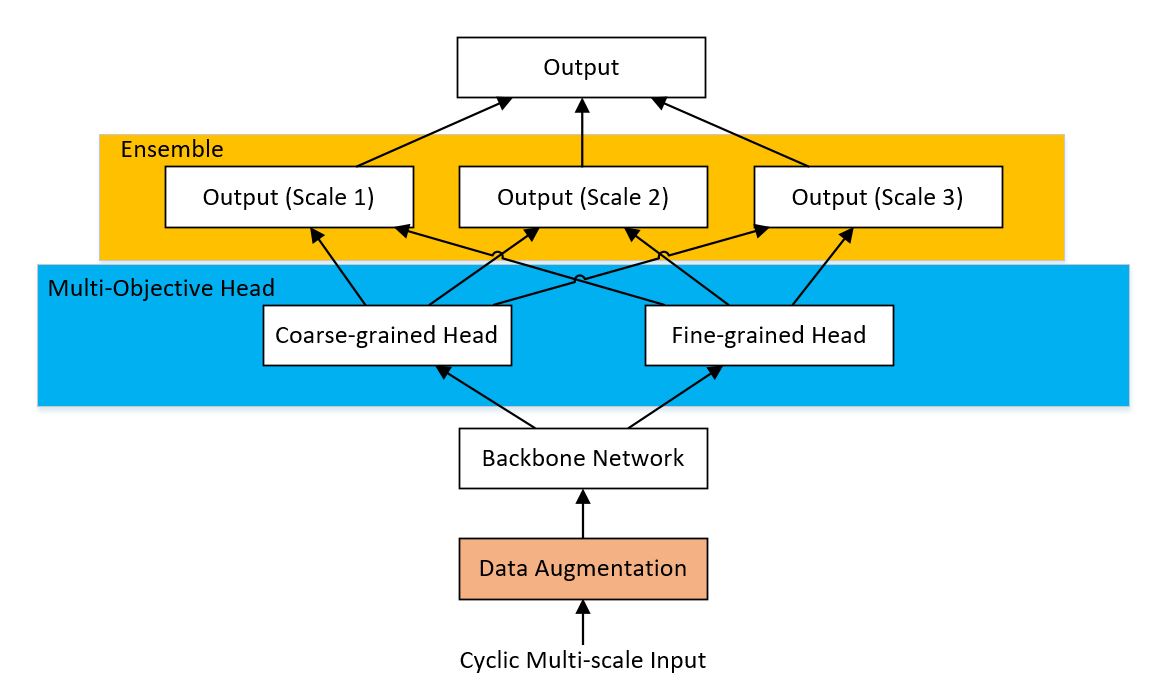}
\caption{Overview of model framework: Input cyclic multi-scale images with data augmentation methods into backbone network to extract features; Extracted Features are fed into multi-objective head to extract coarse-grained and fine-grained logits; Logits are ensembled through TTA methods to output the final result of a single model.}
\label{fig:net}
\end{figure}

Our proposed end-to-end framework is illustrated in Figure~\ref{fig:net}. Firstly, we input multi-scale images based on a cyclic multi-scale training strategy and apply data augmentations to increase the diversity of training data. Then we adopt efficient and light-weight networks (e.g. eca-nfnet-l0~\cite{brock2021high}) as backbones to extract features and training with our designed multi-objective head, which can capture coarse-grained and fine-grained information simultaneously. Finally, during inference stage, we merge logits of different scales and design weighted Top-5 voting to ensemble different models.
% Then we choose efficient and light-weight backbone network to obtain features. Furthermore, we propose multi-objective framework to capture coarse-grained and fine-grained information. Finally, we adopt inference strategies to merge logits of different scales and design weighted Top-5 voting to ensemble different models.

\subsection{Multi-objective Framework Design}
\begin{figure*}
\centering
\begin{minipage}[b]{0.30\linewidth}
\includegraphics[width=0.9\linewidth, height=0.9\linewidth]{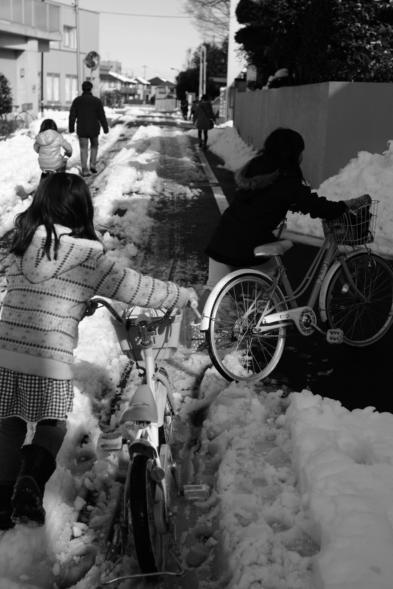}
\end{minipage}
\begin{minipage}[b]{0.30\linewidth}
\includegraphics[width=0.9\linewidth, height=0.9\linewidth]{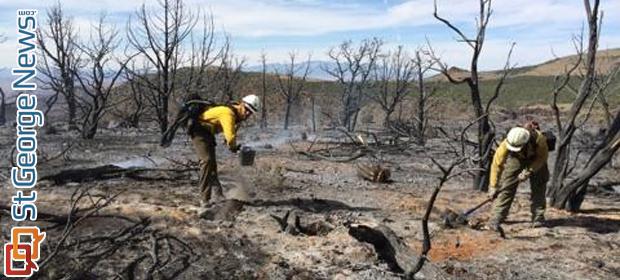}
\end{minipage}
\begin{minipage}[b]{0.30\linewidth}
\includegraphics[width=0.9\linewidth, height=0.9\linewidth]{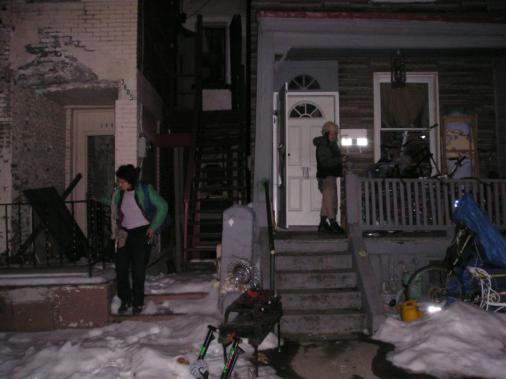}
\end{minipage}
\caption{Examples of bad cases which are easily misclassified as ridiculous results.}
\label{fig:badcase}
\end{figure*}

To capture multi-level semantic information in images, we propose a multi-objective framework. Firstly, domain labels provided by NICO++ dataset naturally contain coarse-grained information and we have considered using them as auxiliary targets to train the backbone network. However, it worsens the performance probably because domain labels focus on the context of images, which may impair the feature learning of foreground objects. Furthermore, we analyse many bad cases, examples from which are illustrated in Figure~\ref{fig:badcase}. We find that bicycle is misclassified as horse, wheat is misclassified as monkey and bicycle is misclassified as gun, respectively, which are far from the correct answer. Therefore, we aim to introduce coarse-grained information to assist model training. Specifically, we manually divide 60 categories into 4 coarse categories according to their properties, denoted as plant, animal, vehicle and object as coarse semantic labels and design a coarse classifier to enable model to learn coarse-grained features. The output dimension is the number of coarse categories, 4, while the output dimension of fine-grained classifier is the number of classes, 60. Under this circumstances, our network can utilize various information from multi-objective, thus increasing robustness and generalization ability of backbone network with barely no computational consumption. 

Besides, we also have explored self-supervised objective to increase the generalization ability of models. However, due to large GPU memory consumption and little improvement on test accuracy, we leave detailed self-supervised task design as future work in domain generalization.

\subsection{Data Augmentation}
Data Augmentation is one of the most significant series of methods in domain generalization, for its simplicity but effectiveness on increasing diversity of data. During training stage, except for common augmentations such as resized-crop, horizontal-flip and normalization, we perform multiple combinations of augmentations and find that the combination of Random Augmentation~\cite{cubuk2020randaugment}, Cutmix~\cite{yun2019cutmix} and Mixup~\cite{zhang2017mixup} and Label Smoothing~\cite{szegedy2016rethinking} is the most effective one.
%  \begin{enumerate}
%  \item  Randomly crop a region whose aspect ratio is in (3/4, 4/3), area is in (0.8, 1.0) and finally resized to the given size.
%  \item Horizontally flip with probability of 0.5.
%  \item Apply random augmentation to decrease computational burden.
%  \item Normalize with mean [0.485, 0.456, 0.406] and standard deviation [0.229, 0.224, 0.225].
%  \item Apply Cutmix and Mixup to form new samples.
%  \item Apply Label Smoothing to avoid overfitting.
%  \end{enumerate}
\subsubsection{Random Augmentation~\cite{cubuk2020randaugment}}
Random Augmentation aims to solve the problem of Auto Augmentation for its high computational cost for the separate search phase on a proxy task, which may be sub-optimal for divergent model and dataset size. It reduces search space for data augmentation and propose one with only 2 hyper-parameters. In this challenge, we set magnitudes, the strength of transformation to 9 and the mean standard deviation to 0.5.
\subsubsection{Cutmix~\cite{yun2019cutmix} and Mixup~\cite{zhang2017mixup}}
Cutmix randomly selects two training images and cuts them into patches with the scale of $\sqrt{1-\gamma}$ on height and width. Then patches from one sample are pasted to another while labels are transformed into one-hot format and mixed proportionally to the area of patches. Mixup also randomly selects two training images and mix them pixel-wise and label-wise with a random number $\lambda$. Both $\gamma$ and $\lambda$ are random numbers, calculated from Beta distribution. In this challenge, we apply these two methods on all batches, with an alternative probability of 0.5. Also, we set $\gamma$ and $\lambda$ to 0.4 and 0.4 by empirical practice, where $\gamma \in\mathbb[0,1]$ and $\lambda \in\mathbb[0,1]$. 
\subsubsection{Label Smoothing~\cite{szegedy2016rethinking}}
Hard label is prone to overfitting practically in deep learning based approaches, and label smoothing was first proposed to change the ground truth label probability to,
\begin{equation}
p_{j} = \left\{
\begin{aligned}
&1-\epsilon, &if j = y_{j},\\
&\epsilon/(N-1), &otherwise.\\
\end{aligned}
\right.
\end{equation}

where $\epsilon$ is a constant, $N$ denotes the number of classes, $j$ is the index of class, $y_{j}$ is the index of ground truth for current image. In this challenge, we set $\epsilon$ to 0.1, where $\epsilon \in\mathbb[0,1)$.
\subsubsection{Others}
Except the above methods, we have also exerted Gaussian Blur, Random Erasing and Image-cut, but fail to improve on public test set probably because of conflicts and overlaps between augmentations. For example, Image-cut is a data extension method to cut original images into five images offline, containing four corners and a center one, which has similar effects with multi-scale training and five-crop. Random Erasing may conflict with Cutmix and Mixup for introducing noise on augmented images and Gaussian Blur may impair the quality of images especially with small objects. 

\subsection{Training Strategy}
Different training strategies may lead to severe fluctuations in deep learning based models. In this section, we propose innovative and effective training strategies to enhance the process of model training.
\subsubsection{Cyclic Multi-scale Training}
Due to various scales of objects in NICO++ dataset, we employ Cyclic Multi-scale Training strategy to increase the robustness and generalization ability of our model. Different from multi-scale strategy in object detection which applies multi-scale input in each batch, we propose to change the input size of data periodically for every 5 epochs to better learn representations of objects at different scales, which is suitable for models without pre-training and consume less GPU memory. Also, as we figure out that larger scale is helpful to improve model performance, we set large multi-scales for light-weight eca-nfnet-l0, and small multi-scales for the rest of backbone networks.
\subsubsection{Others}
Considering the constraints of GPU memory, we adopt gradient accumulation~\cite{ruder2016overview} to increase batch size, which calculates the gradient of a single batch and accumulate for several steps before the update of network parameters and zero-reset of gradient. Besides, we have also verified two-stage training strategies, which CAM~\cite{zhou2016learning} is utilized to extract foreground region during second-stage to fine-tune the model. However, little improvement on test set with longer fine-tuning epochs is not worthy.

\subsection{Inference Strategy}
Inference strategies consume little computational resources but may increase model performance significantly with proper design. In this section, we will introduce our multi-scale inference strategy and weighted Top-5 voting ensemble method. 
% employ two classical strategies with best settings to our empirical knowledge.
\subsubsection{Test-Time Augmentation}
Test-Time Augmentation (TTA) aims to enhance images in test set with proper data augmentation methods and enable models to make predictions on different augmented images to improve model performance. Typical TTA methods including resize, crop, flip, color jitter are used in this challenge, where we first use resize with an extension of 64 pixels on input size. Then we apply different crop strategies, five-crop with an additional extension of 32 pixels, center-crop with an additional extension of 64 pixels. Besides, for center-crop based TTA we use horizontal flip with a probability of 0.5, color jitter with a scope of 0.4, and conduct fused TTA methods based on above. Also, we design multi-scale logits ensemble strategy for multi-scale test. Specifically, we input three different size corresponding to different networks, and apply average-weighted (AW) and softmax-weighted (SW), two different ensemble methods to fuse logits of three scales, as shown in Eq.~\ref{weighted1} and Eq.~\ref{weighted2}, respectively. Finally, we compare different TTA combinations to get the best strategy and remove TTAs contradicting with previous strategies (e.g. five-crop and Image-cut)
% \begin{equation}
%     L_{ens} = \left\{
%     \begin{aligned}
%     &[L_{1}, L_{2}, L_{3}] * [1/3, 1/3, 1/3]^T, &AW,\\
%     &[L_{1}, L_{2}, L_{3}] * Softmax(Max(L_{1}), Max(L_{2}), Max(L_{3}))^T, &SW.\\
%     \end{aligned}
%     \right.
% \label{weighted}
% \end{equation}
\begin{equation}
    L_{AW} = [L_{1}, L_{2}, L_{3}] * [1/3, 1/3, 1/3]^T
\label{weighted1}
\end{equation}
\begin{equation}
    L_{SW} = [L_{1}, L_{2}, L_{3}] * Softmax(Max(L_{1}), Max(L_{2}), Max(L_{3}))^T
\label{weighted2}
\end{equation}

where $L_{AW}$ denotes the ensemble logits by average-weighted method, $L_{SW}$ denotes the ensemble logits by softmax-weighted method, $L_{i}$ denotes the logits of $i$-th scale after applying TTA methods.

\subsubsection{Model Ensemble}
As diverse model may capture different semantic information due to its unique architecture, model ensemble methods are used to better utilize different context of models to make improvement. Logits ensemble and voting are mainstream methods for their simplicity and efficiency. In this challenge, we propose weighted Top-5 voting strategy on diverse models. Specifically, we get the Top-5 class predictions of each model and then assign voting weights for each prediction according to its rank. The voting weights for the top-5 predictions of each model can be formulated as Eq.~\ref{voting}, 
% set weights according to its rank, then sum the weights with identical index and chooses index with maximum weights as the final prediction, as illustrated in , 
% \begin{equation}
% \begin{split}
%     &W_{j} = 1/j, \\
%     &W_{I_{i,j}} = \sum_{I_{i,j}}W_{j},\\
%     &I_{max} = Max(W_{I_{i,j}})
%     \label{voting}
% \end{split}
% \end{equation}
\begin{equation}
W_{Top5} = [1, 1/2, 1/3, 1/4, 1/5]
\label{voting}
\end{equation}
% where $W_{j}$ denotes the index weight of jth position, $I_{i,j}$ denotes the jth class index of ith model, $W_{I_{i,j}}$ denotes sum of weights of $I_{i,j}$ index, and $I_{max}$ denotes the final prediction.
where $W_{Top5}$ denotes the voting weights from 1-st to 5-th. While voting, we sum the voting weights for the same class prediction from different models and finally take the class prediction with the maximum sum of voting weights as the final result. 
\section{Experiments}
\subsection{Implementation Details}
Models are trained on 8 Nvidia V100 GPUs, using AdamW optimizer with cosine annealing scheduler. Learning rate is initialized to $1e^{-3}$ for 300 epochs and weight decay is $1e^{-3}$ for all models. Batch size is 8 and gradient accumulation is adopted to restrict GPU memory.
\subsection{Results}
Three backbone networks, eca-nfnet-l0, eca-nfnet-l2 and efficientnet-b4 are trained with cyclic multi-scale training strategy to enrich the diversity of models for better ensemble results. Data augmentations including Cutmix and Mixup, Random Augmentation and Label Smoothing are adopted with empirical hyper-parameters to get diverse training data. With inference strategies of TTA and model ensemble, including multi-scale logits ensemble, five-crop and weighted Top-5 voting, we further improve test set performance. During phase 1 on public test set, the evaluation metric is Top-1 accuracy, and finally we rank $1^{st}$ with a result of 88.16$\%$. The results are shown in Table~\ref{tab:final}. 
\setlength{\tabcolsep}{4pt}
\begin{table}
\begin{center}
\begin{tabular}{ccc}
\toprule
    Model & Input size & Top-1 Accuracy\\
        \midrule
        eca-nfnet-l0 & Multi-scale(large) & 86.87$\%$\\
        eca-nfnet-l2 & Multi-scale(small) & 86.55$\%$\\
        efficientnet-b4 & Multi-scale(small) & 81.43$\%$\\
        \midrule
    \textbf{ensemble} &  & \textbf{88.16$\%$}\\
\bottomrule
\end{tabular}
\end{center}
\caption{Top-1 accuracy of models on public test set. Multi-scale(small) indicates input size as (448, 384, 320), Multi-scale(large) indicates input size as (768, 640, 512).}
\label{tab:final}
\end{table}
\setlength{\tabcolsep}{1.4pt}

\subsection{Ablation Studies}
We conduct ablation studies to demonstrate the effectiveness of our methods on multi-objective framework design, data augmentations, training and inference strategies, illustrated in Table~\ref{tab:ablation}. These methods are added to Baseline step-by-step and effectively improve performance on public test set with negligible computational resources. For example, Cutmix and Mixup improves 12.19$\%$,  Multi-scale(small) improves 7.72$\%$ and Multi-scale(large) further improves 2.55$\%$. Except the above methods, other strategies basically improve performance for around 2$\%$ without any mutual conflicts.

% Besides, Image-cut can further improve accuracy of 0.1$\%$ with weighted Top-5 voting strategy, probably because local features are addressed to enhance the learning of small-object when ensembled, but performs worse individually because global features are cut-out.
Besides, as mentioned above in Section 4.3, Image-cut has similar effects with multi-scale training and five-crop. Thus, when applying them together, Image-cut can only further improve accuracy of 0.1$\%$ with weighted Top-5 voting strategy. CAM based approach can improve accuracy of 0.4$\%$ but it requires second-stage training, consuming extra 60 epochs. Therefore, we exclude it from our framework for simplicity. However, the local feature view of Image-cut and the object-sensitive features of CAM based methods are still worth to be explored in future research. 
% These methods are theoretically reasonable and may provide further research direction in domain generalization.

Furthermore, we apply several recent state-of-the-art domain generalization methods, including CORAL, SWAD and StableNet, but they decrease the performance by 0.98$\%$, 2.41$\%$, 1.24$\%$ respectively. It further demonstrates that existing algorithms on domain generalization may only benefit on certain dataset and perform worse than heuristic data augmentations.
\begin{table}
\begin{center}
      \begin{tabular}{cc}
        \toprule
        Methods & Top-1 Accuracy\\
        \midrule
        Baseline & 56.61$\%$\\
        +Multi-scale(small) & 64.33$\%$\\
        +Cutmix and Mixup & 76.52$\%$\\
        +Random Augmentation & 78.92$\%$\\
        +Test-time Augmentation & 80.53$\%$\\
        +Multi-objective Framework & 81.53$\%$\\
        +Multi-scale(large) - Multi-scale(small)& 84.08$\%$\\
        +Longer Epochs & 86.87$\%$\\
        \textbf{+Model Ensemble} & \textbf{88.16$\%$}\\
        \midrule
      \end{tabular}
  \caption{Ablation studies on different strategies. Baseline indicates a classic eca-nfnet-l0 backbone network. Except for Model Ensemble, the backbone network of all other strategies is eca-nfnet-l0, and $+$ denotes adding the method based on the previous experimental settings, while $-$ denotes removing the method from the previous experimental settings.}
  \label{tab:ablation}
  \end{center}
 \end{table}

\section{Conclusions}
In this paper, we comprehensively analyse bag of tricks to tackle image recognition on domain generalization. Methods including multi-objective framework design, data augmentations, training and inference strategies are shown to be effective with negligible extra computational resources. By exerting these methods in a proper way to avoid mutual conflicts, our end-to-end framework consumes low-memory usage, but largely increases robustness and generalization ability, which achieves a significantly high accuracy of 88.16$\%$ on public test set and 75.65$\%$ on private test set, and ranks $1^{st}$ in domain generalization task of nicochallenge-2022.

\section{Acknowledgments}
This work is supported by Chinese National Natural Science Foundation (62076033), and The Key R$\&$D Program of Yunnan Province (202102AE09001902-2).

\clearpage
\bibliographystyle{splncs04}
\bibliography{egbib}
\end{document}